\documentclass[twoside]{article}

\usepackage[bindingoffset=0cm,a4paper,centering,textheight=200mm,textwidth=120mm,includefoot,includehead]{geometry}
\usepackage[all]{xy}
\usepackage{tikz}
\usepackage{float}
\usepackage{euscript}
\usepackage{natbib}
\usepackage{epstopdf}
\usepackage{chapterbib}
\usepackage{textcomp}
\usepackage{times}
\usepackage{amssymb}
\usepackage{amsfonts}
\usepackage{latexsym}
\usepackage{url}
\usepackage{epsfig}
\usepackage{amsmath}
\usepackage{amsfonts}
\usepackage{graphics}
\usepackage{graphicx}
\usepackage{makeidx}
\usepackage{psfrag}
\usepackage{array}
\usepackage{tabularx}
\usepackage{xspace}
\usepackage{wrapfig}

\usepackage{times}
\usepackage[scaled]{helvet}
\usepackage{courier}

\usepackage{color}
\usepackage{booktabs}
\usepackage{subfig}

\usepackage{multirow}

\usepackage{ifpdf}

\usepackage{paralist}
\usepackage{enumerate}

\newcommand{\ignore}[1]{}

\makeatother
\begin{document}

\newcommand{\flpmod}{\models_{\mathit{flp}}}
\newcommand{\dual}[1]{\overline{#1}}
\newcommand{\htmod}{\models_{ht}}
\newcommand{\sppmod}{\models_{spp}}
\newcommand{\flp}[2]{#1^{\underline{#2}}}
\newcommand{\gl}[2]{#1^{{#2}}}
\newcommand{\spp}[2]{#1^{\underline{\underline{#2}}}}

\newcommand{\la}{\langle}
\newcommand{\ra}{\rangle}
\newcommand{\st}{\,|\;}
\newcommand{\cA}{\mathcal{A}}
\newcommand{\cT}{\mathcal{T}}
\newcommand{\cH}{\mathcal{H}}
\newcommand{\cF}{\mathcal{F}}
\newcommand{\cMM}{\mathcal{MM}}
\newcommand{\cM}{\mathcal{M}}
\newcommand{\cAS}{\mathcal{AS}}
\newcommand{\cFLP}{\mathcal{FLP}}
\newcommand{\cSP}{\mathcal{SP}}
\newcommand{\cG}{\mathcal{G}}
\newcommand{\cC}{\mathcal{C}}
\newcommand{\cL}{\mathcal{L}}
\newcommand{\cP}{\mathcal{P}}
\newcommand{\vph}{\varphi}
\newcommand{\dom}{\mathit{dom}}
\newcommand{\supp}{\mathit{supp}}
\newcommand{\At}{\mathit{At}}
\newcommand{\hd}{\mathit{hd}}
\newcommand{\bd}{\mathit{bd}}
\newcommand{\tr}{\mathbf{t}}
\newcommand{\fa}{\mathbf{f}}
\newcommand{\un}{\mathbf{u}}
\newcommand{\rra}{\rightarrow}
\newcommand{\Ra}{\Rightarrow}
\newcommand{\La}{\Leftarrow}
\newcommand{\lla}{\leftarrow}
\newcommand{\lra}{\leftrightarrow}
\newcommand{\Lra}{\Leftrightarrow}
\newcommand{\n}{\mathit{not\;}}
\newcommand{\smodels}{\texttt{smodels}}
\newcommand{\lparse}{\texttt{lparse}}

\setcounter{footnote}{0}
\setcounter{section}{0}
\markboth{V.~W.~Marek, I.~Niemel\"a and M.~Truszczy{\'n}ski}{Origins of ASP}
\thispagestyle{empty}

\ \\
\ \\
{\LARGE\bf Origins of Answer-Set Programming -- Some Background And Two 
Personal Accounts}

\ \\
\ \\
\ \\
\ \\
{\large\textbf{Victor W. Marek}}\\
Department of Computer Science\\
University of Kentucky\\
Lexington, KY 40506-0633, USA

\medskip
\noindent
{\large\textbf{Ilkka Niemel\"a}}\\
Department of Information and Computer Science\\
Aalto University\\
Finland

\medskip
\noindent
{\large\textbf{Miros{\l}aw Truszczy{\'n}ski}}\\
Department of Computer Science\\
University of Kentucky\\
USA

\medskip
\ \\
\noindent
{\textbf{Abstract:} 
We discuss the evolution of aspects of nonmonotonic reasoning towards 
the computational paradigm of answer-set programming (ASP). We give a 
general overview of the roots of ASP and follow up with the personal
perspective on research developments that helped verbalize the main 
principles of ASP and differentiated it from the classical logic 
programming.
}

\section{Introduction --- Answer-Set Programming Now}

Merely ten years since the term was first used and its meaning
formally elaborated, answer-set programming has reached the status of
a household name, at least in the logic programming and knowledge 
representation communities. In this paper, we present our personal
perspective on influences and ideas --- most of which can be traced 
back to research in knowledge representation, especially
nonmonotonic reasoning, logic programming with negation, constraint
satisfaction and satisfiability testing --- that led to the two papers 
\cite{mt99,nie99} marking the beginning of answer-set programming as 
a computational paradigm. 

\emph{Answer-set programming} (\emph{ASP}, for short) is a paradigm for 
declarative programming aimed at solving search problems and their
optimization variants. Speaking informally, in ASP a search problem is 
modeled as a theory in some language of logic. This representation is 
designed so that once appended with an encoding of a particular instance
of the problem, it results in a theory whose \emph{models}, under the 
semantics of the formalism, correspond to solutions to the problem for 
this instance. The paradigm was first formulated in these terms by  
\citet{mt99} and \citet{nie99}.

The ASP paradigm is most widely used with the formalism of logic 
programming without function symbols, with programs interpreted by the
\emph{stable-model}
semantics introduced by \citet{gl88}. Sometimes the
syntax of programs is extended with the \emph{strong} negation operator 
and disjunctions of literals are allowed in the heads of program rules. 
The semantics for such programs was also defined by 
\citet{gl90b}. They proposed to use the term \emph{answer sets} for sets
of literals, by which programs in the extended syntax were to be 
interpreted. Ten years after the answer-set semantics was introduced, 
answer sets lent their name to the budding paradigm. However, there is 
more to answer-set programming than logic programming with the 
stable-model and answer-set semantics. Answer-set programming languages 
rooted directly in first-order logic, extending it in some simple 
intuitive ways to model definitions, have also been proposed over the 
years and have just matured to be computationally competitive with the 
original logic programming embodiments of the paradigm 
\citep*{den98,dt08,et04a}.

Unlike Prolog-like logic programming, ASP is fully declarative. Neither
the order of rules in a program nor the order of literals in rules have
any effect on the semantics and only negligible (if any) effect on the 
computation. All ASP formalisms come with the functionality to model
definitions and, most importantly, inductive definitions, in intuitive
and concise ways. Further, there is a growing body of works that start 
addressing methods of modular program design 
\citep*{Dao-TranEFK09,JanhunenOTW09} 
and program development and debugging 
\citep*{BrainV05,BrummayerJ10}. 
These features facilitate modeling problems in ASP, and make 
ASP an approach accessible to non-experts.

Most importantly, though, ASP comes with fast software for 
processing answer-set programs. Processing of programs in ASP is most 
often done in two steps. The first step consists of \emph{grounding} the 
program to its equivalent propositional version. In the second step, 
this propositional program is \emph{solved} by a backtracking search 
algorithm that finds one or more of its  answer sets
(they represent solutions) or determines that no 
answer sets (solutions) exist. The current 
software tools employed in each step, commonly referred to as 
\emph{grounders} and \emph{solvers}, respectively, have already reached
the level of performance that makes it possible to use them successfully
with programs arising from problems of practical importance.

This effectiveness of answer-set programming tools is a result of a 
long, sustained and systematic effort of a large segment of the 
Knowledge Representation
community, and can be attributed to a handful of crucial ideas, some 
of them creatively adapted to ASP from other fields. Specifically,
\emph{domain restriction} was essential to help control the size of 
ground programs. It was implemented in \emph{lparse}, the first 
ASP grounder \cite{ns96}. The \emph{well-founded semantics}
\cite{vrs91} inspired strong propagation methods implemented 
in the first full-fledged ASP solver \emph{smodels} \cite{ns96}. 
\emph{Program completion} \cite{cl78} provided a bridge to satisfiability 
testing. For the class of tight programs \cite{el03}, it allowed for a direct
use of satisfiability testing software in ASP, the idea first implemented
in an early version of the solver \emph{cmodels}\footnote{\url{http://www.cs.utexas.edu/users/tag/cmodels.html}}. \emph{Loop 
formulas} \cite{lz02} extended the connection to satisfiability testing to 
arbitrary programs. They gave rise to such successful ASP solvers as 
\emph{assat} \cite{lz02}, \emph{pbmodels} \cite{lt05b} and later 
implementations of \emph{cmodels} \cite{cmodels-2}. Database techniques 
for \emph{query optimization} influenced the design of the grounder for 
the \emph{dlv} system\footnote{\url{www.dbai.tuwien.ac.at/proj/dlv/}}
\citep*{dlvtocl06}. Important advances of satisfiability testing 
including the data structure of \emph{watched literals}, \emph{restarts},
and \emph{conflict-clause learning} were incorporated into the ASP solver 
\emph{clasp}\footnote{\url{www.cs.uni-potsdam.de/clasp/}}, at present 
the front-runner among ASP solvers and the winner of one track of the 
2009 SAT competition. Some of the credit for the advent of high-performance 
ASP tools is due to the initiative to hold ASP grounder and solver 
contests. The two editions of the contest so far
\citep*{asp-c07,DeneckerVBGT09} focused on modeling 
and on solver performance, and introduced a necessary competitive element 
into the process. 

The modeling features of ASP and computational performance of ASP 
software find the most important reflection in a growing range of
successful applications of ASP. They include molecular biology 
\citep*{GebserGISSTV10,GebserKSTV10}, decision support system for space 
shuttle controllers \citep*{BalducciniGN06}, phylogenetic systematics 
\citep*{Erdem11}, automated music composition \citep*{Boenn2011tplp},
product configuration \citep*{SoininenN99,TSNS2003:iced,fs11}
and repair of web-service workflows \citep*{FriedrichFMPT10}.

And so, ASP is now a declarative programming paradigm built on top of a 
solid theoretical foundation, with features that facilitate its use in 
modeling, with software supporting effective computation, and with a 
growing list of successful applications to its credit. How did it all 
come about? This paper is an attempt to reconstruct our personal journey
to ASP.

\section{Knowledge Representation Roots of Answer-Set Programming}

One of the key questions for knowledge representation is how to model
commonsense knowledge and how to automate commonsense reasoning. The 
question does not seem particularly relevant to ASP understood, as it 
now commonly is, as a general purpose computational paradigm for solving
search problems. But in fact, knowledge representation research was 
essential. First, it recognized and emphasized the importance of 
principled modeling of commonsense and domain knowledge. The impact of
the modeling aspect of knowledge representation and reasoning is distinctly 
visible in the current implementations of ASP. They support high level
programming that separates modeling problem specifications from problem
instances, provide intuitive means to model aggregates, and offer direct
means to model defaults and inductive definitions. Second, knowledge
representation research, and especially nonmonotonic reasoning research,
provided the theoretical basis for ASP formalisms: the answer-set 
semantics of programs can be traced back to the semantics
of default logic and autoepistemic logic, the semantics of the logic
FO(ID) \citep*{den00,dt08}  has it roots in the well-founded semantics of
nonmonotonic provability operators. 

In this section we discuss the development of those ideas in knowledge 
representation that eventually took shape of answer-set programming. 
In their celebrated 1969 paper, McCarthy and Hayes wrote
\begin{quote}
\emph{\mbox{[...]} intelligence has two parts, which we shall call the
epistemological and the heuristic. The epistemological part is the
representation of the world in such a form that the solution of problems
follows from the facts expressed in the representation. The heuristic
part is the mechanism that on the basis of the information solves the
problem and decides what to do.}
\end{quote}
With this paragraph McCarthy and Hayes ushered knowledge 
representation and reasoning into artificial intelligence and moved 
it to one of the most prominent positions in the field. Indeed, what
they
referred to as the \emph{epistemological part} is now understood as
knowledge representation, while the \emph{heuristic part} has evolved 
into broadly understood automated reasoning --- a search for proofs or 
models.

The question how to do knowledge representation and reasoning quickly
reached the forefront of artificial intelligence research. McCarthy
suggested first-order logic as the formalism for knowledge
representation. The reasons behind the proposal were quite appealing.
First-order logic is ``descriptively universal'' and proved itself
as the formal language of mathematics. Moreover, key reasoning tasks in 
first-order logic could be automated, assuming one adopted appropriate
restrictions to escape semi-decidability of first-order logic in
its general form.

However, there is no free lunch and it turned out that
first-order logic 
could not be just taken off the shelf and used for knowledge 
representation with no extra effort required. The problem is that 
domain knowledge is rarely complete. More often than not, information 
available to us has gaps. And the same is true for artificial agents 
we would like to function autonomously on our behalf. Reasoning with 
incomplete knowledge is inherently \emph{defeasible}. Depending on how 
the world turns out to be (or depending on how the gaps in our knowledge
 are closed), some conclusions reached earlier may have to be withdrawn.
The monotonicity of first-order logic consequence relation is at odds with 
the \emph{nonmonotonicity} of defeasible reasoning and makes modeling 
defeasible reasoning in first-order logic difficult.

To be effective even when available information is incomplete, 
humans often develop and use \emph{defaults}, that is, rules that 
typically work but in some exceptional situations should not be used.
We are good at learning defaults and recognizing situations in which
they should not be used. In everyday life, it is thanks to defaults
that we are not bogged down in the \emph{qualification problem} 
\cite{mca77}, that is, normally we do not check that every possible 
precondition for an action holds before we take it. And we naturally 
take advantage of the \emph{frame axiom} \cite{mh69} when reasoning, 
that is, we take it that things 
remain as they are unless they are changed by an action. Moreover, 
we do so avoiding the difficulties posed by the \emph{ramification 
problem} \cite{finger87}, which is concerned with side effects of actions. However,
first-order logic conspicuously lacks defaults in its syntactic 
repertoire nor does it provide an obvious way to simulate them. It is not 
at all surprising, given that defaults have a defeasible flavor about
them. Not being aware that a situation is ``exceptional'' one may
apply a default but later be forced to withdraw the conclusion upon
finding out the assumption of ``non-exceptionality'' was wrong.

Yet another problem for the use of first-order logic in knowledge
representation comes from the need to model definitions, most notably
the inductive ones. 
The way humans represent definitions has an aspect of defeasibility 
that is related to the \emph{closed-world assumption}. Indeed, we 
often define a concept by specifying all its known instantiations. 
We understand such a definition as meaning also that \emph{nothing else}
is an instance of the concept, even though we rarely if ever say it 
explicitly. But the main problem with definitions lies elsewhere. 
Definitions often are \emph{inductive} and their correct meaning is 
captured by the notion of a least fixpoint. First-order logic cannot 
express the notion of a least fixpoint and so does not provide a way 
to specify inductive definitions.

These problems did not go unrecognized and in late 1970s researchers 
were seeking ways to address them. Some proposals called for extensions
of first-order logic by explicit means to model defaults while other 
argued that the language can stay the same but the semantics had to 
change. In 1980, the Artificial Intelligence Journal published a double 
issue dedicated to nonmonotonic reasoning, a form of reasoning based 
on but departing in major ways from that in first-order logic. The 
issue contained three papers by \citet{mca80}, \citet{re80}, and 
\citet{mcdd80}
that launched the field of nonmonotonic reasoning.

McCarthy's proposal to bend the language of first-order logic to the
needs of knowledge representation was 
to adjust the semantics of first-order logic and
to base the entailment relation 
among sentences in first-order logic on \emph{minimal} models only
\cite{mca80}. He called the resulting formalism \emph{circumscription} 
and demonstrated how circumscription could be used in several settings 
where first-order logic failed to work well. \citet{re80} extended
the syntax of first-order logic by \emph{defaults}, inference rules with 
exceptions, and described formally reasoning with defaults. Reiter was 
predominantly interested in reasoning with \emph{normal} defaults but 
his default logic was much more general. Finally, McDermott and Doyle 
proposed a logic based on the language of modal logic which, as they 
suggested, was also an attempt to model reasoning with defaults. This 
last paper was found to suffer from minor technical problems. Two years 
later, \citet{mcd82} published another paper which corrected and 
extended the earlier one.

These three papers demonstrated that shortcomings of first-order logic 
in modeling incomplete knowledge and supporting reasoning from these 
representations could be addressed without giving up on the logic entirely
but by adjusting it. They sparked a flurry of research activity directed 
at understanding and formalizing nonmonotonic reasoning. One of the most
important and lasting outcomes of those efforts was the autoepistemic 
logic proposed by \citet{mo84,mo85}. Papers by Moore can be regarded
as closing the first phase of the nonmonotonic reasoning as a field of
study. 


Identifying nonmonotonic reasoning as a phenomenon deserving an in-depth
study was a major milestone in logic, philosophy and artificial
intelligence. The prospect of understanding and automating reasoning 
with incomplete information, of the type we humans are so good at, 
excited these research communities and attracted many researchers to 
the field. Accordingly, the first 10-12 years of nonmonotonic reasoning 
research brought many fundamental results and established solid 
theoretical foundations for circumscription \cite{mca80,li88}, default 
logic \citep*{rc81,hmd86,mt89a,pea90}, autoepistemic logic 
\citep*{mo85,nie88a,mt88,shv90,shv90a} 
and modal nonmonotonic logics in the style of McDermott and Doyle 
\citep*{mst91a,sch91b,st92}. Researchers made
progress in clarifying the relationship between these 
formalisms 
\cite{ko87,koerratum,mt89c,bf91b,tru90b}. Computational aspects received
much attention, too. First 
complexity results appeared in late 1980s and early 1990s 
\cite{CL90,mt88,ks89,got92,Stillman92} and 
early, still naive at that time, implementations of automated reasoning 
with nonmonotonic logics were developed around the same time 
\cite{et87,NT86:hlm,NT87:hlm,gin89}. 
Several research monographs were published in late 1980s and early 
1990s systematizing that phase of nonmonotonic reasoning research and 
making it accessible to outside communities \cite{bes89,brb91,mt93}. 

Expectations brought up by the advent of nonmonotonic reasoning 
formalisms were high. It was thought that nonmonotonic logics
would facilitate concise and elaboration tolerant representations of 
knowledge, and that through the use of defeasible inference rules like
defaults it would support fast reasoning. However, around the time of 
the first Knowledge Representation and Reasoning Conference, KR 1989 in 
Toronto, concerns started to surface in discussions and papers. 

First, there was the issue of multiple belief sets, depending on the
logic used represented as extensions or expansions. A prevalent 
interpretation of the problem was that multiple belief sets provided
the basis for \emph{skeptical} and \emph{brave} modes of reasoning. 
Skeptical reasoning meant considering as consequences a reasoner was
sanctioned to draw only those formulas that were in every belief set. 
Brave reasoning required a non-deterministic commitment to one of the 
possible belief sets with all its elements becoming consequences of the
underlying theory (in the nonmonotonic logic at hand). The first 
approach was easy to understand and accept at the intuitive level. But 
as a reasoning mechanism it was rather weak as in general it supported 
few non-trivial inferences. The second approach was underspecified ---
it provided no guidelines on how to select a belief set, and it was
not at all obvious how if at all humans perform such a selection. 
Both skeptical and brave reasoning suffered from the fact that there 
were no practical problems lying around that could offer some direction
as to how to proceed with any of these two approaches. 

Second, none of the main nonmonotonic logics seemed to provide a good
formalization of the notion of a default or of a defeasible consequence
relation. This was quite a surprising and in the same time worrisome
observation. Nonmonotonic reasoning brought attention to the concept
of default and soon researchers raised the question of how to reason
\emph{about} defaults rather than with defaults \citep*{pea90,klm90,lm92}. A
somewhat different version of the same question asked about defeasible 
consequence relations, whether they can be characterized in terms of 
intuitively acceptable axioms, and whether they have semantic 
characterizations \cite{gab85,mak88}. Despite the success of 
circumscription, default and autoepistemic logics in addressing several
problems of knowledge representation, it was not clear if or how they 
could contribute to the questions above. In fact, it still remains
an open problem whether any deep connection between these logics and
the studies of abstract nonmonotonic inference relations exists.
 
Next, the complexity results obtained at about same time
\cite{mt88,eg91,got92,Stillman92,eg93,eg95}
were viewed as negative. They dispelled 
any hope of higher computational efficiency of nonmonotonic reasoning. 
Even under the restriction to the propositional case, basic reasoning 
tasks turned out to be as complex as and in some cases even more 
complex (assuming polynomial hierarchy does not collapse) than reasoning
in propositional logic. Even more discouraging results were obtained 
for the general language.

Finally, the questions of applications and implementations was becoming 
more and more urgent. There were no practical artificial intelligence 
applications under development at that time that required nonmonotonic 
reasoning. Nonmonotonic logics continued to be extensively studied and 
discussed at AI and KR conferences, but the belief that they can have 
practical impact was diminishing. There was a growing feeling that they
might amount to not much more but a theoretical exercise. Complexity 
results notwithstanding, the ultimate test of whether an approach is 
practical can only come from experiments, as the worst-case 
complexity is one thing but real life is another. But there was little
work on implementations and one of the main reasons was lack of test 
cases whose hardness one could control. Researchers continued to 
analyze ``by hand'' small examples arguing about correctness of their 
default or autoepistemic logic representations. These toy examples were
appropriate for the task of understanding basic reasoning patterns. But
they were simply too easy to provide any meaningful insights into 
automated reasoning algorithms and their performance.

And so the early 1990s saw a growing sentiment that in order to prove itself,
to make any lasting impact on the theory and practice of knowledge 
representation and, more generally, on artificial intelligence, 
practical and efficient systems for nonmonotonic reasoning had to be
developed and their usefulness in a broad range of applications
demonstrated. Despite of all the doom and gloom of that time, there
were reasons for optimism, too. The theoretical understanding of 
nonmonotonic logics reached the level when development of sophisticated 
computational methods became possible. Complexity results were 
disappointing but the community recognized that they concerned the 
worst case setting only. 
Human experience tells us that there are good reasons to think that real life
does not give rise to worst-case instances too often, in fact, that it 
rarely does. Thus, through experiments and the focus on reasoning with
structured theories one could hope to obtain efficiency sufficient for 
practical applications. Moreover, it was highly likely that once 
implemented systems started showing up, they would excite the community,
demonstrate the potential of nonmonotonic logics, and spawn competition 
which would result in improvements of algorithms and performance
advances. 

It is interesting to note that many of the objections and criticisms 
aimed at nonmonotonic reasoning were instrumental in helping to identify
key aspects of answer-set programming. Default logic did not provide
an acceptable formalization of reasoning about defaults but inspired 
the answer-set semantics of logic programs \cite{bf87a,gl88,gl90b} and
helped to solve a long-standing problem of how to interpret negation in 
logic programming. Answer-set programming, which adopted the syntax of 
logic programs, as well as the answer-set semantics, can be regarded as
an implementation of a significant fragment of default logic. The lack 
of obvious test cases for experimentation with implementations forced 
researchers to seek them 
outside of artificial intelligence and led them to the area of graph 
problems. This experience showed that the phenomenon of multiple belief 
sets can be turned from a bug to a feature, when researchers realized 
that it allows one to model \emph{arbitrary} search problems, with 
extensions, expansions or answer sets, depending on the logic used,
representing problem solutions \citep*{ceg97,mr03}. 

However important, knowledge representation was not the only source of 
inspiration for ASP. Influences of research in several other areas of 
computer science, such as databases, logic programming and satisfiability, 
are also easily identifiable and must be mentioned, if only
briefly. 
One of the key themes in research in logic programming in the 1970s and 1980s
was the quest for the meaning of the negation operator. Standard 
logic programming is built around the idea of a single intended Herbrand 
model. A program represents the declarative knowledge about the domain
of a problem to solve. Some elements of the model, more accurately, ground 
terms the model determines, represent solutions to the problem. All works
well for Horn programs, with the least Herbrand model of a Horn program 
as the natural choice for the intended model. But the negation operator,
being ingrained in the way humans describe knowledge, cannot be avoided.
The logic programming community recognized this and the negation was an element of 
Prolog, an implementation of logic programming, right from the very 
beginning. And so, the question arose for a declarative (as opposed to 
the procedural) account of its semantics. 

Subsequent studies identified a non-classical nature of the negation 
operator. This nonmonotonic aspect of the negation operator in logic 
programming was also a complicating factor in the effort to find a 
single intended model of logic programs with negation. It became clear 
that to succeed one either had to restrict the class of programs or 
to move to the three-valued settings. The first line of research 
resulted in an important class of stratified programs \citep*{abw87}, 
the second one led \citet{fi85} and \citet{kun87} to the 
\emph{Kripke-Kleene model} and, later on, 
\citet*{vrs91} to the well-founded model.

In the hindsight, the connection to knowledge representation and 
nonmonotonic reasoning should have been quite evident. However, the
knowledge representation and logic programming communities had little
overlap at the time. And so it was not before the work by 
\citet{bf87a} and \citet{ge87} that the connection was 
made explicit and then exploited. That work demonstrated that intuitive 
constraints on an intended model cannot be reconciled with the 
requirement of its uniqueness. In other words, with negation in the
syntax, we must accept the reality of multiple intended models.
The connection between logic programming and knowledge representation,
especially, default and autoepistemic logics was important. On the one
hand, it showed that logic programming can provide syntax for an 
interesting non-trivial fragment of these logics, and drew attention of 
researchers attempting implementations of nonmonotonic reasoning systems.
On the other hand, it led to the notion of a stable model of a logic
program with negation. It also reinforced the importance of the key 
question how to adapt the phenomenon of multiple intended models for 
problems solving.

The work in databases provided a link between query languages and logic
programming. One of the outcomes of this work was $\mathrm{DATALOG}$, 
a fragment of logic programming without function symbols, proposed as a
query language. The database research resulted in important theoretical 
studies concerning complexity, expressive power and connection of  
$\mathrm{DATALOG}$ to the SQL query language \citep{ceg97}. 
$\mathrm{DATALOG}$ was implemented, for instance as a part of DB2 
database management system. $\mathrm{DATALOG}$ introduced an important 
distinction between extensional and intentional database components. 
Extensional database is the collection of tables that are stored in the 
database, the corresponding relation names known as extensional 
predicate symbols. The intensional database is a collection of 
intentional tables defined by $\mathrm{DATALOG}$ queries. In time this 
distinction was adopted by
answer-set programming as a way to separate problem specification from 
data. The database community also considered extensions of $\mathrm{DATALOG}$
with the negation connective. Because of the semantics of the resulting 
language, 
multiplicity of answers in $\mathrm{DATALOG}^\neg$ was a problem, as it was in
a more general setting of arbitrary programs with negation. Therefore, 
$\mathrm{DATALOG}^\neg$ never turned into a practical 
database query language
(although, its stratified version could very well be used to this end).
However, it was certainly an interesting fragment of logic programming.
And even though its expressive power was much lower than that of general
programs,\footnote{It has to be noted though that the expressive power 
of general programs with function symbols and negation
goes well beyond what could be accepted 
as computable under all reasonable semantics \citep*{sch91,mnr94}.} 
there was hope that fast tools to process 
$\mathrm{DATALOG}^\neg$ can be developed. Jumping ahead, we note here
that it was $\mathrm{DATALOG}^\neg$ that was eventually adopted as the
basic language of answer-set programming. 



\section{Towards Answer-Set Programming at the University of Kentucky}

Having outlined some of the key ideas behind the emergence of answer-set
programming, we now move on to a more personal account of research ideas 
that eventually resulted in the formulation of the answer-set programming
paradigm. In this section, Victor Marek and Mirek Truszczynski, discuss
the evolution of their understanding of nonmonotonic logics and how they 
could be used for computation that led to their paper \emph{Stable logic 
programming --- an alternative logic programming paradigm} \cite{mt99}. 
A closely intertwined story of Ilkka Niemel\"a, follows in the subsequent
section. As the two accounts are strongly personal and necessarily quite
subjective, for the most part they are given in the first person. And so, 
in this section ``we'' and us refers to Victor and Mirek, just as ``I'' 
in the next one to Ilkka. 

In mid 1980s, one of us, Victor, started to study nonmonotonic logics 
following a suggestion from Witold Lipski, his former Ph.D. student 
and close collaborator. Lipski drew Victor's attention to Reiter's
papers on closed-world assumption and default logic \cite{re78,re80}. 
In 1984, Victor attended the first Nonmonotonic Reasoning Workshop at 
Mohonk, NY, and came back convinced about the importance of problems 
that were discussed there. In the following year, he attracted Mirek 
to the program of the study of mathematical foundations of nonmonotonic 
reasoning. 

In 1988 Michael Gelfond visited us in Lexington and in his presentation 
talked about the use of autoepistemic logic \cite{mo85} to provide a 
semantics to logic programs. At the time we were already studying
autoepistemic logic, inspired by talks Victor attended at Mohonk and
by Moore's paper on autoepistemic logic in the Artificial Intelligence 
Journal \cite{mo85}. We knew by then that stable sets of formulas of
modal logic, introduced by \citet{st80} and shown to be essential
for autoepistemic logic, can be constructed by an iterated inductive 
definition from their modal-free part \cite{ma86}. We also realized the 
importance of a simple normal form for autoepistemic theories introduced 
by \citet{ko87}. 

Thus, we were excited to see that logic programs can be understood as 
some simple autoepistemic theories thanks to Gelfond's interpretation 
\cite{ge87}. Soon thereafter, we also realized that logic programs could 
be interpreted also as default logic theories and that the meaning of 
logic programs induced on them by default logic extensions is the same as
that induced by autoepistemic expansions \cite{mt89c}. It is important to 
note that default logic was first used to assign the meaning to logic 
programs by \citet{bf87a}, but we did not know 
about their work at the time. Bidoit and Froidevaux effectively defined 
the stable model semantics for logic programs. They did so indirectly and
with explicit references to default extensions. The direct definition of
stable models in logic programming terms came about one year later in the 
celebrated paper by \citet{gl88}. 

What became apparent to us soon after Gelfond's visit was that despite
both autoepistemic expansions and default extensions inducing the same
semantics on logic programs, it was just serendipidity and not the 
result of the inherent equivalence of the two logics. In fact, we 
noticed that there was a deep mismatch between Moore's autoepistemic 
logic with the semantics of expansions and Reiter's default logic with 
the semantics of extensions. In the same time, we discovered a form of 
default logic, to be more precise, an alternative semantics of 
default logic, which was the perfect match for that of expansions for 
autoepistemic logic \cite{mt89a}. This research culminated about 15 
years later with a paper we co-authored with Marc Denecker that provided 
a definitive account of the relationship between default and 
autoepistemic logics \citep*{dmt03} and resolved problems and flaws of 
an earlier attempt at explaining the relationship due to 
\citet{ko87}. Another paper in this volume \citep*{dmt11} discusses the 
informal basis for that work and summarizes all the key results.
  
The relationship between default and autoepistemic logic was of only 
marginal importance for the later emergence of answer-set programming.
But another result inspired by Gelfond's visit turned out to be 
essential. In our study of autoepistemic logic we wanted to establish 
the complexity of the existence of expansions. We obtained the result 
by showing that the problem of the existence of a stable model of a 
logic program is NP-complete and, by doing so, we obtained the same 
complexity for the problem of the existence of expansions of 
autoepistemic theories of some simple form but still rich enough to 
capture logic programs under Gelfond's interpretation \cite{mt88}.

The result for autoepistemic logic did not turn out to be particularly 
significant as the class of autoepistemic
theories it pertained to was narrow. And it
was soon supplanted by a general result due to \citet{got92},
who proved the existence of the expansions problem to be 
$\Sigma_2^P$-complete. But it was an entirely different matter with 
the complexity result concerning the existence of stable models of
programs! 

First, our proof reduced a combinatorial problem, that of the
existence of a kernel in a directed graph, to the existence of stable 
model of a suitably defined program. This was a strong indication that
stable semantics may, in principle, lead to a general purpose formalism
for solving combinatorial and, more generally, search problems. Of 
course we did not fully realize it at the time. Second, it was quite 
clear to us, especially after the first KR conference in Toronto in
May 1989 that the success of nonmonotonic logics can come only with 
implementations. Many participants of the conference (we recall David 
Poole and Matt Ginsberg being especially vocal) called for working 
systems. Since by then we understood the complexity of stable-model 
computation, we asked two University of Kentucky students Elizabeth 
and Eric Freeman to design and implement an algorithm to compute stable 
models of propositional programs. They succeeded albeit with limits 
--- the implementation could process programs with about 20 variables 
only. Still, theirs was most likely the first working implementation 
of stable-model computation. Unfortunately, with the M.S. degrees 
under their belts, Eric and Elizabeth left the University of Kentucky. 

For about three years after this first dab into implementing reasoning
systems based on a nonmonotonic logic, our attention was focused on
more theoretical studies and on the work on a monograph on mathematical
foundations of nonmonotonic reasoning based on the paradigm of 
context-dependent reasoning. However, the matter of implementations
had constantly been on the backs of our minds and in 1992 we decided
to give the matter another try. As we felt we understood default logic 
well and as it was commonly viewed as the nonmonotonic logic of the 
future, in 1992 we started the project, Default Reasoning System DeReS.
We aimed at implementing reasoning in the unrestricted language of
propositional default logic. We also started a side project to DeReS,
the TheoryBase project, aimed at developing a software system generating
default theories to be used for testing DeReS. The time was right as two
promising students, Pawel Cholewinski and Artur Mikitiuk, joined the 
University of Kentucky to pursue doctorate degrees in computer science. 

As is common in such circumstances, we were looking for a sponsor of 
this research and found one in the US Army Research Office (US ARO),
which was willing to support this work. A colleague of ours, Jurek 
Jaromczyk, also at the University of Kentucky, coined the term DeReS, 
a pun on an old polish word ``deresz'' presently rarely used and meaning 
a stallion, quite appropriate for the project to be conducted in 
Lexington, ``the world capital of the horse.'' In the proposal to US ARO
we promised to investigate basic reasoning problems of default logic:
\begin{enumerate}
\item Computing of extensions
\item Skeptical reasoning with default theories --- testing if a formula 
belongs to all extensions of an input default theory
\item Brave reasoning with default theories --- testing if a formula 
belongs to some extension of an input default theory.
\end{enumerate}
The basic computational device was backtracking search for a basis of 
an extension of a finite default theory $(D,W)$. This was based on two
observations due to Reiter: that while default extensions of a finite 
default theory are infinite, they are finitely generated; and that the
generators are all formulas of $W$ and the consequent formulas of some
defaults from $D$. We also employed ideas such as relaxed stratification 
of defaults \cite{cho94,litu94} for pruning the search space and relevance
graphs for simplifying provability.

We also thought it was important to have the nonmonotonic reasoning 
community accept the challenge of developing implementations of automated
nonmonotonic reasoning. Our proposal to US ARO contained a request for 
funding of a retreat dedicated to knowledge representation, nonmonotonic 
reasoning and logic programming. The key goals for the retreat were:
\begin{enumerate}
\item To stimulate applications of nonmonotonic formalisms and
implementations of automated reasoning systems based on nonmonotonic
logics
\item To promote the project to create a public domain library of
benchmark problems in nonmonotonic reasoning.
\end{enumerate}
We held the workshop in Shakertown, KY, in October 1994. Over 30 leading 
researchers in nonmonotonic reasoning participated in talks and 
we presented there early prototypes of DeReS and TheoryBase. 
Importantly, we heard then for the first time from Ilkka Niemel\"a 
about the work on systems to perform nonmonotonic reasoning in the 
language of logic programs in his group at the Helsinki University of 
Technology. The meeting helped to elevate the importance of implementations 
of nonmonotonic reasoning systems and their applications. It evidenced first 
advances in the area of implementations, as well as in the area of 
benchmarks, essential as so far most problems considered as benchmarks 
were toy problems such as ``Tweety'' and ``Nixon Diamond.'' 

The DeReS system was not designed with any specific applications in 
mind. At the time we believed that, since default logic could model
several aspects of commonsense reasoning, once DeReS became available, 
many artificial intelligence and knowledge representation researchers 
would use it in their work. And we simply regarded broadly understood 
knowledge representation problems as the main application area for 
DeReS. 

Working on DeReS immediately brought up to our attention the question 
of testing and performance evaluation. In the summer of 1988, Mirek 
attended a meeting on combinatorics where Donald Knuth talked about 
the problem of testing graph algorithms and his proposal how to do it 
right. Knuth was of the opinion that testing algorithms on randomly
generated graphs is insufficient and, in fact, often irrelevant. Graphs
arising in real-life settings rarely resemble graphs generated at random
from some probabilistic model. To address the problem, Knuth developed
a software system, Stanford GraphBase, providing a mechanism for 
creating collections of graphs that could be then used in projects
developing graph algorithms. Graphs produced by the Stanford GraphBase 
were mostly generated from real-life objects such as maps, dictionaries,
novels and images. Some were based on rather obscure sources such as 
sporting events in Australia. The documentation was superb (the book by
Knuth on the Stanford GraphBase is still available). The Stanford GraphBase
was free 
and its use was not restricted. From our perspective, two aspects were 
essential. First, the Stanford GraphBase provided a unique identifier to
every graph it created and so experiments could be described in a way
allowing others to repeat them literally and perform comparisons on 
identical sets of graphs. Second, the Stanford GraphBase supported 
creating families of examples similar but increasing in size, thus 
allowing to test scalability of algorithms being developed.

In retrospect, the moment we started talking about testing our 
implementations of default logic was the defining moment on our path
towards the answer-set programming paradigm. Based on our complexity
result concerning the existence of stable models and its implication for
default logic, we knew that all NP-complete graph problems could be 
reduced to the problem of the existence of extensions. The reductions
expressed instances of graph problems as default theories. Thus, in order
to get a family of default theories, similar but growing in size, we
needed to select an NP-complete problem on graphs (say, the hamiltonian
cycle problem), generate a family of graphs, and generate for each graph
in the family the corresponding default theory. These theories could be
used to test algorithms for computing extensions. This realization gave
rise to the TheoryBase, a software system generating default theories 
based on reductions of graph problems to the existence of the extension 
problem and developed on top of the Stanford GraphBase, which served as
the source of graphs. The TheoryBase provided default theories based on
six well-known graph problems: the existence of $k$-colorings, Hamiltonian
cycles, kernels, independent sets of size at least $k$, and vertex covers
of size at most $k$. As the Stanford GraphBase provided an unlimited 
supply of graphs, the TheoryBase offered an unlimited supply of default 
theories.

We will recall here the TheoryBase encoding of the existence of a 
$k$-coloring problem as it shows that already then some fundamental
aspects of the methodology of representing search problems as default
theories started to emerge. Let $G = (V,E)$ be an undirected graph 
with the set of vertices $V = \{v_1,\ldots,v_n\}$. Let $C = \{c_1,\ldots,
c_k\}$ be a set of colors. To express the property that vertex $v$ is
colored with $c$, we introduced propositional atoms $clrd(v,c)$. For 
each vertex $v_i, i=1,\ldots, n$, and for each color $c_j, j=1,\ldots, 
k$, we defined the default rule
\[
color(v_i,c_j) = \frac{: \neg clr(v_i,c_1),..., \neg clr(v_i,c_{j-1}),
 \neg clr(v_i,c_{j+1}),..., \neg clr(v_i,c_k)}{clr(v_i,c_j)}.
\]
The set of default rules $\{color(v_i,c_j) : j=1,\ldots,k\}$ models
a constraint that vertex $v_i$ obtains exactly one color. The default
theory $(D_0, \emptyset)$, where 
\[
D_0 = \{color(v_i,c_j) : i=1,\dots,n,\ \ j=1,\ldots,k\},
\]
has $k^n$ extensions corresponding to all possible colorings (not 
necessarily \emph{proper}) of the vertices of $G$. Thus, the default
theory $(D_0, \emptyset)$ defines the basic space of objects within 
which we need to search for solutions. In the present-day 
answer-set programming implementations choice or cardinality rules, 
which offer much more concise representations, are used for that 
purpose. Next, our TheoryBase encoding imposed constraints to eliminate 
those colorings that are not proper. To this end, we used additional
default rules, which we called {\em killing} defaults, and which now
are typically modeled by logic program rules with the empty head.
To describe them we used a new propositional variable $F$ 
and defined
\[
\mathit{local}(e,c) = \frac{clrd(x,c) \wedge clrd(y,c):\neg F}{F},
\]
for each edge 
$e=(x,y)$ 
of the graph and for each color $c$.
Each default $\mathit{local}(e,c)$ ``kills'' all color assignments which give
color $c$ to both ends of edge $e$. It is easy to check (and it also
follows from now well-known more general results) that defaults
of the form $\mathit{local}(e,c)$ ``kill'' all non-proper colorings and leave
precisely those that are proper. This two-step modeling methodology,
in which we first define the space of objects that contains all solutions,
and then impose constraints to weed away those that fail some problem
specifications, constitutes the main way by which search problems are 
modeled in ASP. 

The key lesson for us from the TheoryBase project was that combinatorial
problems can be represented as default theories and that constructing 
these representations is easy. It was then for the first time that we 
sensed that programs finding extensions of default theories could be
used as general purpose problem solving tools. It also lead us, in our 
internal discussions to thinking about ``second-order'' flavor of 
default logic, given the way it was used for computation. Indeed, in all theories
we developed for the TheoryBase, extensions rather than their single 
elements represented solutions. In other words, the main reasoning
task did not seem to be that of skeptical or brave reasoning (does a 
formula follow skeptically or bravely from a default theory) but 
computing \emph{entire} extensions. We talked about this second-order
flavor when presenting our paper on DeReS at the KR conference in 1996
\citep*{cmt96}. At that time, we knew we were closing in on a new declarative
problem-solving paradigm based on nonmonotonic logics.

A problem for us was, however, a fairly poor performance of DeReS. 
The default extensions are closed under consequence. This means that 
processing of default theories requires testing provability of 
prerequisites and justifications of defaults. This turned out to be 
a major problem affecting the processing time of our implementations.
It is not surprising at all in view of the complexity results of
\citet{got92} and \citet{Stillman92}. Specifically, existence of extensions
is a $\Sigma_2^P$-complete problem.

There is, of course, an easy case of provability when all formulas in
a default theory are conjunctions of literals only. Now the problem 
with the provability of premises disappears. However, DeReS organized
its search for solutions by looking for sets of generating defaults,
inheriting this approach from the case of general default theories, 
rather than for literals generating an extension. And that was still a 
problem. There are typically many more rules in a default theory than
atoms in the language.

At the International Joint Conference and Symposium on Logic Programming
in 1996, Ilkka and his student Patrik Simons presented the first report 
on their \emph{smodels} system \cite{ns96}. But it seems fair to
say that only a similar presentation and a demo Ilkka gave at the Logic
Programming and 
Non-Monotonic Reasoning Conference in 1997, in Dagstuhl, made the 
community really take notice. The \emph{lparse/smodels}
constituted a major conceptual breakthrough and handled nicely all the 
traps DeReS did not avoid. First, \emph{lparse/smodels} focused on the 
right fragment of default logic, logic programming with the stable-model 
semantics. Next, it organized search for a stable model by looking for atoms
that form it.  Finally, it supported programs with variables and separated, as
was
the standard in logic programming and databases, a program (a problem
specification) from an extensional database (an instance of the problem).

The work by Niemel\"a had us focus our thinking about nonmonotonic
logics as computational devices on the narrower but all-important case
of logic programs. We formulated our ideas about the second-order
flavor of problem solving with nonmonotonic logics and contrasted them
with the traditional Prolog-style interpretation of logic programming.
We stated our initial thoughts on the methodology of problem solving
that exploited our ideas of modeling combinatorial problems that we
used in the TheoryBase project, as well as the notion of program-data 
separation that came from the database community and was, as we 
just mentioned, already used in our field by Niemel\"a. These ideas
formed the backbone of our paper on an alternative way logic programming
could be used for solving search problems \cite{mt99}.

\section{Towards Answer-Set Programming at the Helsinki University of
Technology} 

In this section Ilkka Niemel\"a discusses the developments at the Helsinki
University of Technology that led to the paper \emph{Logic Programs with Stable
Model Semantics as a Constraint Programming Paradigm}~\cite{nie99}.
Similarly as in the previous section, the account is very personal and quite
subjective. Hence, in this section "I" refers to Ilkka. 

I got exposed to nonmonotonic reasoning when I joined the group of Professor
Leo
Ojala at the Helsinki University of Technology in 1985. The group was studying
specification and verification techniques of distributed systems. One of the
themes was specification of distributed systems using modal, in particular,
temporal and dynamic logics. The group had got interested in the solutions of
the \emph{frame problem} based on nonmonotonic logics when looking for compact
and computationally efficient logic-based specification techniques for
distributed systems. My role as a new research assistant in the group was to
examine autoepistemic logic by Moore, nonmonotonic modal logics by McDermott
and Doyle, and default logic by Reiter from this perspective. 

There was a need for tool support and together with a doctoral student Heikki
Tuominen we developed a system that we called the Helsinki Logic Machine, ``an
experimental reasoning system designed to provide assistance needed for
application oriented research in logic''~\citep{NT86:hlm,NT87:hlm}. The system
included tools for theorem proving, model synthesis, model checking, formula
manipulation for modal, temporal, epistemic, deontic, dynamic, and
\emph{nonmonotonic} logics. It was written in Quintus Prolog and 
contained implementations, for instance, for Reiter's default logic, McDermott
and Doyle style nonmonotonic modal logic, and autoepistemic logic \emph{in the
propositional case} based on the literature and some own
work~\citep{et87,mcdd80,nie88a}. While nonmonotonic reasoning was a side-track
in the Helsinki Logic Machine, it seems that it was one of the earliest
working nonmonotonic reasoning systems although we were not very well aware
of this at the time. 

The work and, in particular, the difficulties in developing efficient tools led
me to further investigations to gain a deeper understanding of algorithmic
issues and related complexity
questions~\citep*{nie88a,Niemela88:csl,Niemela90:jelia,nie92}.
Similar questions were studied by others and in the early 90s results
explaining the algorithmic difficulties started emerging. These results showed
that key reasoning tasks in major nonmonotonic logics are complete for the
second level of the polynomial
hierarchy~\cite{CL90,got92,Stillman92,nie92}.  This indicated that
these nonmonotonic logics have \emph{two orthogonal sources of
  complexity} that we called classical reasoning and conflict resolution. 
Orthogonality means that even if we assume that classical reasoning
can be done efficiently, nonmonotonic reasoning still remains
NP-hard (unless the polynomial hierarchy collapses). 
 
These results made me to focus more on conflict resolution to develop
techniques for pruning the search space of potential expansions/extensions. One
approach was to develop compact characterizations of expansions/extensions
capturing their key ingredients. For autoepistemic logic I developed such a
characterization based on the idea that expansions can be captured in terms of
the modal subformulas in the premises and classical reasoning and exploited the
idea in a decision procedure for autoepistemic logic~\cite{nie88a}.
Together with Jussi Rintanen we also showed that if one limits the theory in
such a way that conflict resolution is easy by requiring stratification, then
efficient reasoning is possible by further restrictions affecting the other
source of complexity~\cite{NR92:kr}. 

The characterization based on modal subformulas generalizes also to default
logic where extensions can be captured using justifications in the rules and
leads to an interesting way of organizing the search for expansions/extensions
as a \emph{binary search tree} very similar to that in the DPLL algorithm for
SAT~\cite{Niemela94:kr,Niemela95:ijcai}. Further pruning techniques can be
integrated to cut substantial parts of the potential search space for
expansions/extensions and exploit, for instance, stratified parts of the rule set.  My
initial but very unsystematic experimentation gave promising results. 

In 1994 encouraged and challenged by the Shakertown Workshop organized
by Victor and Mirek, I decided to restrict to a simple subclass of
default theories, that is, logic programs with the stable model
semantics. For this subclass classical reasoning is essentially limited
to Horn clauses and can be done efficiently in linear time using
techniques proposed by Dowling and Gallier in the
1980s~\cite{dg84}.  I had no particular application in mind. The
goal was to study whether the conflict resolution techniques I had
developed for autoepistemic and default logic would scale up so that it
would be possible to handle \emph{very large sets of rules} which meant
at that time thousands or even tens of thousands of rules.

At that time Patrik Simons joined my group and started working on a C++
implementation of the general algorithm tailored to logic
programs. Patrik had excellent insights into the key implementation
issues from very early on and the first version was released in
1995~\cite{NS95:ijcaiws}. The C++ implementation was 
called \smodels{} and it computed stable models for ground normal programs. It
gave surprising good results immediately and could handle programs with a few
thousand ground rules. Challenge benchmarks were combinatorial problems, mainly
colorability and Hamiltonian cycles, an idea that I learnt from Mirek and
Victor in Shakertown. 
For such hard problems the performance of \smodels{} was substantially 
better than state-of-the-art tools such as the SLG system~\cite{ChenW96}.
 
When developing benchmarks for evaluating novel algorithmic ideas and
implementation techniques we soon realized that working with ground
programs is too cumbersome. In practice, for producing large enough
interesting ground programs for benchmarking we needed to write separate
programs in some other language to generate ground logic programs. This
took considerable time for each benchmark family and was quite
inflexible and error-prone. We realized that in order to attract users
and to be able to attack real applications we needed to support logic
program rules with variables.

For handling rules with variables we decided to employ a two level 
architecture. The first phase was concerned with \emph{grounding}, a
process to generate a set of ground instances of the rules in the program
so that stable models are preserved. Actual stable-model computation was
taking place in the second \emph{model search} phase on the program 
grounded in the previous one. The idea was to have a separation of concern, 
that is, 
be able to exploit advanced database and other such techniques in the 
first phase and novel search and pruning techniques in the other in such 
a way that both steps could be developed relatively independently. We 
released the first such system in 1996~\cite{ns96}. 

This was a major step forward in attracting users and getting closer to
applications.  Such a system supporting rules with variables enabled
\emph{compact and modular} encodings of problems without any further host
language. It was now also possible to separate the problem specification and
the data providing the instance to be solved. 

Working with the system and studying potential applications made me realize
that logic programming with the stable model semantics is very different from
traditional logic programming implemented in various Prolog systems. These
systems are answering queries by SLD resolution and producing answer
substitutions as results. But we were using logic programs more like in a
constraint programming approach where rules are seen as constraints on a
solution set (stable model) of the program and where a solution is not an
answer substitution but a stable model, that is, a valuation that satisfies all
the rules. This is like in constraint satisfaction problems where a solution is
a variable assignment satisfying all the constraints. I wrote down these ideas
in a paper \emph{Logic Programs with Stable Model Semantics as a Constraint
Programming Paradigm} which was first presented in a workshop on Computational
Aspects of Nonmonotonic Reasoning in 1998~\cite{Niemela98:cnmr} and then
appeared as an extended journal version in 1999~\cite{nie99}. The paper
emphasized, in particular, the knowledge representation advantages of logic
programs as a constraint satisfaction framework:
\begin{quote}
``Logic programming with the stable model semantics is put forward as an
  interesting constraint programming paradigm.  It is shown that the
  paradigm embeds classical logical satisfiability but seems to provide
  a more expressive framework from a \emph{knowledge representation} point of
  view.''
\end{quote}

In 1998 we put more and more emphasis on potential applications and, in
particular, on product configuration. This made us realize that a more
efficient grounder supporting an extended modeling language is
needed. At that point another student, Tommi Syrj\"anen, with excellent
implementation skills and insight on language design, joined the group
and work on a new grounder, \lparse{}, started. The goal was to enforce
a tighter typing of the variables in the rules to facilitate the
application of more advanced database techniques for grounding and the
integration of built-in predicates and functions, for instance, for
arithmetic.
 
We also realized that for many applications normal logic programs were
inadequate not allowing compact and intuitive encodings. This led to the
introduction of new language constructs: (i)~choice rules for encoding choices
instead of recursive odd loops needed in normal programs and (ii)~cardinality
and weight constraints for typical conditions needed in many practical
applications~\cite{SoininenN99,NSS99:lpnmr}. In order to fully exploit the
extensions computationally Patrik Simons developed techniques to provide
built-in support for them also in the model search phase in the version 2 of
\smodels~\cite{Simons99:lpnmr}.

So in 1999 when Vladimir Lifschitz coined the term answer-set programming, the
system that we had with \lparse{} as the grounder and \smodels{} version 2 as
the model search engine offered quite promising performance. For example, for
propositional satisfiability the performance of \smodels{} compared nicely to
the best SAT solvers at that time (before more efficient conflict driven
clause learning solvers like zchaff emerged).  Moreover, very interesting
serious application work started. For example, at the Helsinki University of
Technology we cooperated with the product data management group on automated
product configuration which eventually led to a spin-off company Variantum
(\url{http://www.variantum.com/}). Moreover, in Vienna the dlv project for
handling disjunctive programs had started a couple years earlier and had
already made promising progress. 

\section{Conclusions}

Now, more than 12 years since ASP became a recognizable paradigm of
search problem solving, we see that the efforts of researchers in various 
domains: artificial intelligence, knowledge representation, nonmonotonic 
reasoning, satisfiability and others resulted in a programming formalism 
that is being used in a variety of areas, but principally in those
where the modelers face the issues of defaults, frame axioms and other 
nonmonotonic phenomena. The experience of ASP programmers shows
that these phenomena can be naturally incorporated into the practice of
modeling real-life problems.

We believe ASP is here to stay. It provides a venue for problem
modeling, problem description and problem solving. This does not mean that 
the process of developing ASP is finished. Certainly new extensions of 
ASP will emerge in the future. Additional desiderata include: software 
engineering tools for testing correctness of implementation, integrated 
development environments and other tools that will speed up the process 
of the use of ASP in normal programming practice.  
Better grounders and better solvers able to work with incremental 
grounding only will certainly emerge.
Similarly, new application domains will surface and bring new generations of
investigators and, more importantly, users for ASP.

\section*{Acknowledgments}

The work of the second author was partially supported by the Academy of Finland
(project 122399).  The work of the third author was partially supported by the
NSF grant IIS-0913459.

{\small

}

\end{document}